\documentclass{article}
\usepackage{spconf,amsmath,graphicx}
\usepackage{amsfonts}
\usepackage{multirow}
\usepackage{multicol}
\usepackage{makecell}
\usepackage{array}
\usepackage{xcolor}
\usepackage{caption}
\usepackage{subcaption}
\usepackage{CJKutf8}
\usepackage{hyperref}
\hypersetup{hidelinks}

\newcolumntype{P}[1]{>{\centering\arraybackslash}p{#1}}
\newcolumntype{M}[1]{>{\centering\arraybackslash}m{#1}}

\title{A context-aware knowledge transferring strategy for CTC-based ASR}
\name{Ke-Han Lu, and Kuan-Yu Chen}
\address{National Taiwan University of Science and Technology, Taiwan \\ \texttt{khlu@nlp.csie.ntust.edu.tw,  kychen@mail.ntust.edu.tw}}
\begin{document}
\begin{CJK}{UTF8}{gkai}

%\ninept
%
\maketitle
\begin{abstract}
Non-autoregressive automatic speech recognition (ASR) modeling has received increasing attention recently because of its fast decoding speed and superior performance. Among representatives, methods based on the connectionist temporal classification (CTC) are still a dominating stream. However, the theoretically inherent flaw, the assumption of independence between tokens, creates a performance barrier for the school of works. To mitigate the challenge, we propose a context-aware knowledge transferring strategy, consisting of a knowledge transferring module and a context-aware training strategy, for CTC-based ASR. The former is designed to distill linguistic information from a pre-trained language model, and the latter is framed to modulate the limitations caused by the conditional independence assumption. As a result, a knowledge-injected context-aware CTC-based ASR built upon the wav2vec2.0 is presented in this paper. A series of experiments on the AISHELL-1 and AISHELL-2 datasets demonstrate the effectiveness of the proposed method.
\end{abstract}
\section{Introduction}
\label{sec:intro}

Automatic speech recognition (ASR) systems aim at converting a given input speech signal into its corresponding token sequence. They are not only required to model the acoustic information from the speech signal but needed to generate a precise token sequence corresponding to the speech and the contextual coherence. In recent years, connectionist temporal classification (CTC)-based ASR systems \cite{graves2006connectionist} have attracted significant attention since they can achieve a much faster decoding speed in the non-autoregressive manner and obtain competitive or even better performance compared to the conventional auto-regressive models \cite{ 9414594, nozaki21_interspeech, higuchi2022hierarchical, fujita2022multi, 9413806, 9747887, 9688157}. To be specific, a standard CTC-based ASR usually consists of a multi-layer Transformer-based acoustic encoder and a classification head based on some layers of simple feedforward neural network. The acoustic encoder concentrates on encapsulating important characteristics of input speech into a set of feature vectors. The classification head aims at translating the set of feature vectors into a sequence of tokens. Subsequently, a CTC loss is employed to guild the model training so as to minimize the differences between the generated token sequence and the target gold reference.

Although CTC-based ASR models demonstrate their efficiency and effectiveness on several benchmark corpora, the school of models usually suffers from the conditional independence assumption, making it difficult to consider the relationships among tokens occurring in a sequence. Take the wav2vec2.0-CTC model (cf. Section \ref{sec:wav2vec2-ctc}) as an example, we found most of the substitution errors are mistakenly predicted to tokens with similar pronunciations. We also observed that the output representations generated by the encoder for tokens with similar pronunciations usually mix together. These observations indicate that CTC-based ASR systems can learn acoustic information well but are still imperfect in learning linguistic information.

Various research has been devoted to improving CTC-based ASR. The intermediate CTC \cite{9414594} introduces auxiliary CTC losses to the intermediate layers of the acoustic encoder. The self-conditioned CTC \cite{nozaki21_interspeech} and its variants \cite{higuchi2022hierarchical,fujita2022multi} take predictions from intermediate layers as additional clues for the following layers of the encoder. These methods mainly focus on easing the conditional independence assumption from a theoretical perspective. The contextualized CTC loss \cite{9413806} is proposed to guild the model learn contextualized information by introducing extra prediction heads to predict surrounding tokens. Some studies aspire to improve CTC-based ASR via knowledge transferring from pre-trained language models \cite{9747887}.

Following the research line, in this study, we present a knowledge-injected context-aware CTC-based ASR built upon the wav2vec2.0 \cite{baevski2020wav2vec}. Specific characteristics are at least threefold. First, a knowledge transferring module is designed to distill linguistic information from a pre-trained language model and inject the knowledge into the ASR model. Next, a context-aware training strategy is proposed to relax the conditional independence assumption. Finally, to enjoy the merits of a pre-trained speech representation learning method, the wav2vec2.0 \cite{baevski2020wav2vec} is employed as the acoustic encoder for our ASR model. As a result, a context-aware knowledge transferred wav2vec2.0-CTC ASR (CAKT) model is proposed in this study. It not only has a similar model size and decoding speed to the vanilla CTC-based ASR model, but also inherits benefits from a pre-trained language model and a speech representation learning method. Fig.\ref{fig:model} depicts the architecture of the CAKT model. Extensive experiments are conducted on the AISHELL-1 and AISHELL-2 datasets, and the CAKT yields about 14\% and 5\% relative improvements over the baseline system, respectively. Furthermore, we will release the pre-trained Mandarin Chinese wav2vec2.0 model, the first publicly available speech representation model trained on the AISHELL-2 dataset, to the community.

\section{Related works}
\label{sec:related_works}
Large-scale pre-trained models have attracted much attention in recent years because they are trained using unlabeled data and achieve superior results on several downstream tasks by simple fine-tuning with only a few task-oriented labeled data. In the context of natural language processing, the pre-trained language models, such as bidirectional encoder representations from Transformers (BERT) \cite{devlin-etal-2019-bert} and its variants\cite{liu2019roberta, yang2019xlnet}, are representatives. These models have demonstrated their achievements in information retrieval, text summarization, and question answering, to name just a few. Because of the success, previous studies have investigated the pre-trained language model to enhance the performance of ASR. On the one hand, several studies directly leverage a pre-trained language model as a portion of the ASR model \cite{9755057, 9688009, 9746316, 9413668, NEURIPS2020_7a6a74cb, zheng-etal-2021-wav-bert, yi2021efficiently}. Although such designs are straightforward, they can obtain satisfactory performances. However, these models often slow down the decoding speed and usually have a large set of model parameters. On the other hand, a school of research makes the ASR model to learn linguistic information from pre-trained language models in a teacher-student training manner \cite{9437636, Futami2020, 9747887, 9664007, 9746801}. These models still obtain a fast decoding speed, but their improvements are usually incremental. 

Apart from natural language processing, self-supervised speech representation learning creates a potential research subject in the speech processing community. Representative models include wav2vec \cite{schneider19_interspeech}, vq-wav2vec \cite{Baevski2020vq-wav2vec}, wav2vec2.0 \cite{baevski2020wav2vec}, Hubert \cite{hsu2021hubert}, and so forth. These methods are usually trained with unlabeled data self-supervised and concentrate on deriving informative acoustic representatives for a given speech. Downstream tasks can be done by simply fine-tuning some additional layers of neural networks with only a few task-oriented labeled data. Several studies have explored novel ways to build ASR systems based on the pre-trained speech representation learning models. The most straightforward method is to employ them as an acoustic feature encoder and then stack a simple layer of neural network on top of the encoder to do speech recognition \cite{baevski2020wav2vec}. After that, some studies present various cascade methods to concatenate pre-trained language and speech representation learning models for ASR \cite{9688009, 9746316, NEURIPS2020_7a6a74cb, zheng-etal-2021-wav-bert}. Although these methods have proven their capabilities and effectiveness on benchmark corpora, their complicated model architectures and/or large-scaled model parameters have usually made them hard to be used in practice. 

\section{Proposed Methodology}
\label{sec:methods}

\subsection{Vanilla wav2vec2.0-CTC ASR Model}
\label{sec:wav2vec2-ctc}
Among the self-supervised speech representation learning methods, wav2vec can be treated as the pioneer study in the research subject. In this study, we thus employ the advanced variant, i.e., wav2vec2.0, to serve as the cornerstone. The wav2vec2.0 consists of a CNN-based feature encoder and a contextualized acoustic representation extractor based on multi-layer Transformers. More formally, it takes a raw speech $\mathbf{X}$ as an input and outputs a set of acoustic representations $\mathbf{H}^\text{X}_L$:
\begin{equation}
\mathbf{H}^\text{X}_0 = \text{CNN}(\mathbf{X}),    
\end{equation}
\begin{equation}
\mathbf{H}^\text{X}_l = \text{Transformer}_l(\mathbf{H}^\text{X}_{l-1}),
\end{equation}
where $l \in \{1, \cdots, L\}$ denotes the layer number of Transformers and $\mathbf{H}^\text{X}_{l} \in \mathbb{R}^{d \times T}$ represents a set of $T$ feature vectors whose dimension is $d$. To construct an ASR model, a layer normalization (LN) layer, a linear layer and a softmax activation function are sequentially stacked on the top of the wav2vec2.0:
\begin{equation}
\label{eq:acoustic_representation}
\mathbf{H}^\text{X} = \text{LN}(\mathbf{H}^\text{X}_{L}),
\end{equation}
\begin{equation}
\hat{\mathbf{Y}} = \text{Softmax}(\text{Linear}(\mathbf{H}^\text{X})).
\end{equation}
Consequently, the CTC loss $\mathcal{L}_\text{CTC}$ is used to guide the model training toward minimizing the differences between the prediction $\hat{\mathbf{Y}}$ and the ground-truth $\mathbf{Y}$. We denote the simple but straightforward wav2vec2.0-CTC ASR model as w2v2-CTC. 

% \subsection{Transfer learning module}
\subsection{Token-dependent Knowledge Transferring Module}
\label{sec:proposed}
Extending from the vanilla w2v2-CTC ASR model, we present a token-dependent knowledge transferring module and a context-aware training strategy to not only copy linguistic information from a pre-trained language model to the ASR model but also reduce the limitations caused by the conditional independence assumption. Since the classic BERT model remains the most popular, we thus use it as an example to conduct the framework.

In order to distill knowledge from BERT, a token-dependent knowledge transferring module, which is mainly based on multi-head attention, is introduced. First of all, for each training speech utterance $\mathbf{X}$, special tokens [BOS] and [EOS] are padded at the beginning and end of its corresponding gold token sequence $\mathbf{Y} = \{y_1, \cdots, y_N \}$. Then, as with the seminal literal, we sum each token embedding with its own absolute sinusoidal positional embedding \cite{vaswani2017attention}, which is used to distinguish the order of each token in the line. The set of resulting vectors $\mathbf{E} = \{e^\text{[BOS]},e^{y_1}, \cdots, e^{y_N},e^\text{[EOS]} \}$ and high-level acoustic representations $\mathbf{H}^\text{X}$ (cf. Eq. (\ref{eq:acoustic_representation})) are passed to a multi-head attention layer together:
\begin{equation}
\mathbf{O} = \text{Multi-head Attention}(\mathbf{E}, \mathbf{H}^\text{X}, \mathbf{H}^\text{X}),
\end{equation}
where text-level feature $\mathbf{E}$ is used to query acoustic-level statistics $\mathbf{H}^\text{X}$, and $\mathbf{O}$ denotes a set of $d$-dimensional output representations $\{o^\text{[BOS]},o^{y_1},$ $\cdots, o^{y_N},o^\text{[EOS]} \}$. By doing so, a set of token-dependent anchors (i.e., $\mathbf{E}$) is created and is used to reorganize and aggregate the high-level acoustic representations $\mathbf{H}^\text{X}$. The multi-head attention is employed to fulfill the cross-modality interaction. Consequently, a set of token-dependent acoustic-level representations $\mathbf{O}$ is derived.

\begin{figure}[t]
    \centering
    \includegraphics[width=0.47\textwidth]{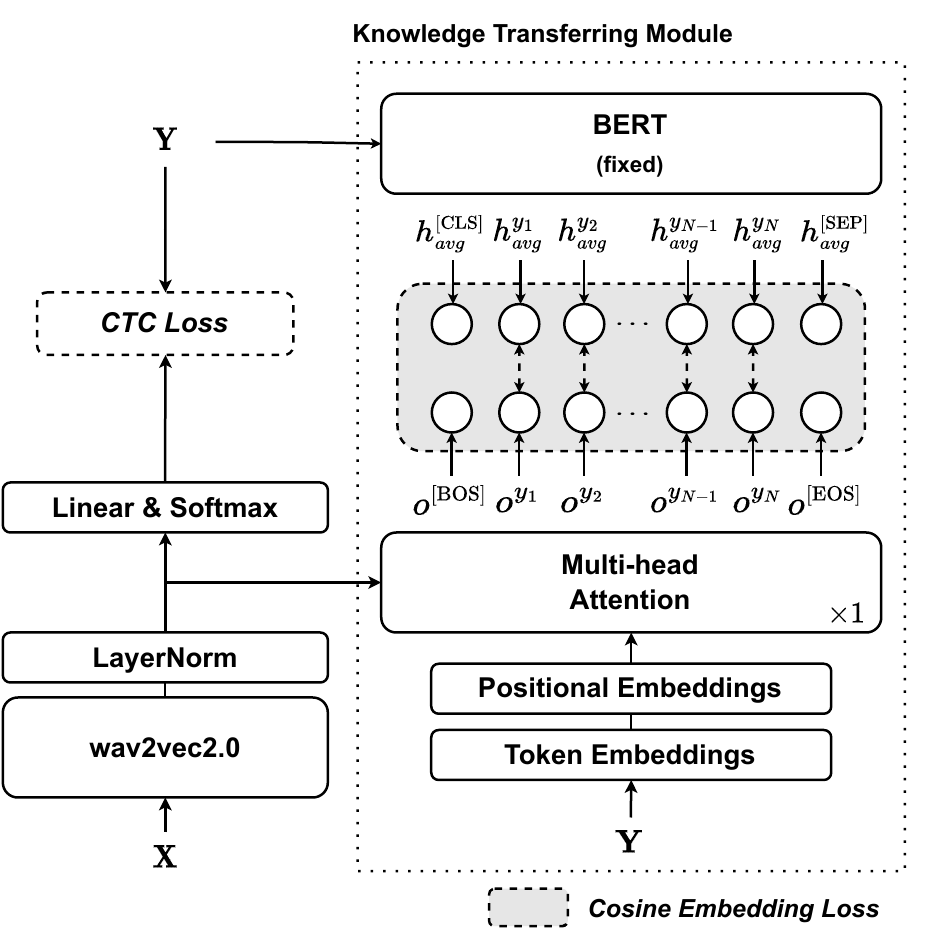}
    \caption{Model architecture of the proposed context-aware knowledge transferred wav2vec2.0-CTC ASR mode.}
    \label{fig:model}
\end{figure}

Previous literal has indicated that each layer of Transformer in BERT encodes different grains of information of a natural language \cite{van2019does,de-vries-etal-2020-whats,derose2020attention}, so we use BERT to encode the target transcription $\mathbf{Y}$ and generate multi-grained contextual representations for each token:
\begin{equation}
\mathbf{H}^\text{Y}_{l} = \{h^\text{[CLS]}_l,h^{y_1}_l, \cdots, h^{y_N}_l,h^\text{[SEP]}_l \} = \text{BERT}_{l}(\mathbf{H}^\text{Y}_{l-1}),
\end{equation}
where $l \in \{1, \cdots, L\}$ denotes the layer number of Transformers in BERT, and $\mathbf{H}^\text{Y}_{0}$ is a set of BERT input vectors converted from $\mathbf{Y}$. We collect all the distilled knowledge from BERT to be the learning target for the token-dependent knowledge transferring module:
\begin{equation}
\mathbf{H}^\text{Y}_{avg} = \text{Avg}(\mathbf{H}^{\text Y}_0, \cdots ,\mathbf{H}^{\text Y}_{L}),
\end{equation}
\begin{equation}
\mathcal{L}_\text{KT} =k\sum^{N}_{n=1} (1 - \text{cos}(\mathbf{h}^{y_n}_{avg}, \mathbf{o}^{y_n})),
\end{equation}
where the objective function $\mathcal{L}_\text{KT}$ is defined to minimize the cosine embedding loss, and a scaling hyper-parameter $k$ is used to equalize the numerical imbalance between the cosine embedding loss and other losses \cite{9747887}. The index $0$ and $N+1$ denote the positions of the special tokens, which are ignored in calculating the training loss. Since the query vectors (i.e., $\mathbf{E}$) equip explicit token information, the multi-head attention can more easily reorganize acoustic features corresponding to each query (token). Consequently, linguistic information can be transferred more efficiently so as to derive a more robust ASR model. The model architecture is depicted in Fig.\ref{fig:model}.

\begin{figure}
    \centering
    \begin{subfigure}[b]{0.23\textwidth}
        \centering
        \includegraphics[width=\textwidth]{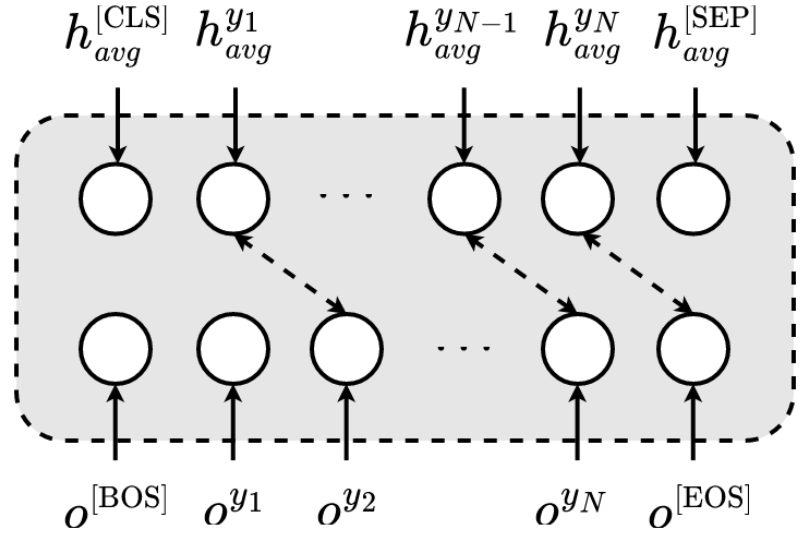}
        \caption{Right-shift Method}
    \end{subfigure}
    \begin{subfigure}[b]{0.23\textwidth}
        \centering
        \includegraphics[width=\textwidth]{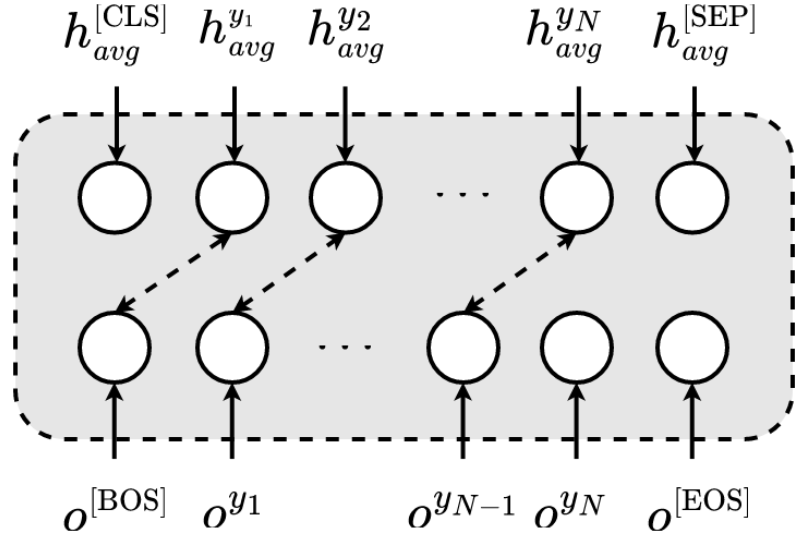}
        \caption{Left-shift Method}
    \end{subfigure}
    % \begin{subfigure}[b]{0.2\textwidth}
    %     \centering
    %     \includegraphics[width=\textwidth]{figures/no_shift.pdf}
    %     \caption{no shift}
    % \end{subfigure}
    \caption{Illustration of right-shift and left-shift methods used in context-aware training strategy. Nodes without aligned are ignored during training.}
    \label{fig:shift}
\end{figure}
 
\subsection{Context-aware Training Strategy}

In order to relax the conditional independence assumption, we present a simple but efficient context-aware training strategy. Our idea is to make the model aware of vicinity information. Hence, instead of distilling knowledge from BERT for each token itself, we force them to learn to predict their adjacency tokens. In order to make the idea work, we shift the source and target representations one position to the left or right, respectively. In formal terms, each pair of $\mathbf{h}^{y_n}_{avg}$ and $\mathbf{o}^{y_{n+1}}$ is aligned together using the right-shift method. Instead, $\mathbf{h}^{y_n}_{avg}$ and $\mathbf{o}^{y_{n-1}}$ become a pair if the left-shift method is used. Fig.\ref{fig:shift} illustrates the idea. Consequently, the knowledge transferring loss $\mathcal{L}_\text{KT}$ becomes:
\begin{equation}
\mathcal{L}_\text{KT} =k\sum^{N}_{n=1} (1 - \text{cos}(\mathbf{h}^{y_n}_{avg}, \mathbf{o}^{y_{n+i}})) 
\left\{ 
  \begin{array}{ c l }
    i=-1 & \textrm{left} \\
    i=1 & \textrm{right} 
  \end{array},
\right.
\end{equation}
where the scaling factor $k$ is empirically set to 20.

During training, the ASR model is trained in a multi-tasking manner and the objective function is to minimize the weighted sum of the classic CTC loss and the cosine embedding loss:
\begin{equation}
\mathcal{L} = \lambda \mathcal{L}_\text{CTC} + (1 - \lambda )\mathcal{L}_\text{KT}.
\end{equation}
The hyper-parameter $\lambda$ is set to 0.3 in our experiments. 

In the inference stage, the knowledge transferring module is discarded and only the CTC branch is used to do speech recognition. Based on the proposed token-dependent knowledge transferring module and context-aware training strategy, the ASR model can not only transfer the linguistic knowledge from a powerful pre-trained language model but also loose the conditional independence assumption. Hence, the resulting ASR model enjoys the benefit from a pre-trained language model without introducing extra model parameters as well as sacrificing the decoding speed. We denote the context-aware knowledge transferred wav2vec2.0-CTC ASR model as CAKT in short hereafter.

\section{Experiment setup}
\label{sec:experiment_setup}
We evaluate the proposed CAKT on two Mandarin Chinese speech corpora, the AISHELL-1 \cite{8384449} and AISHELL-2 \cite{aishell2}. The former is a popular-used benchmark consisting of 178-hour speech data, and the latter is a 1,000-hour industry-scaled speech dataset. Their vocabulary sizes are 4,233 and 7,001, respectively.

We use Fairseq \cite{ott2019fairseq} toolkit to pre-train a Mandarin Chinese wav2vec2.0 on the AISHELL-2 corpus. The CNN-based feature encoder has 512 channels with strides (5,2,2,2,2,2) and kernel sizes (10,3,3,3,3,2,2). The contextualized acoustic representation extractor consists of 12 Transformer layers, each with 12 attention heads, the hidden size is set to 768, and the intermediate size is 3,072. The pre-trained wav2vec2.0 is publicly available at \url{https://github.com/kehanlu/mandarin-wav2vec2}.

Our implementations are conducted on the Espnet2 \cite{watanabe2018espnet} toolkit. The pre-trained wav2vec2.0 is used to initialize the CTC branch. The linear layer, stacked on top of the wav2vec2.0, is token-wisely initialized using BERT token embeddings, except non-verbal tokens. The knowledge transferring module consists of a token embedding layer with 768 dimensions, a sinusoidal absolute positional embedding layer, a multi-head attention layer with 12 heads and 768 model dimensions, and a pre-trained BERT\footnote[1]{\url{https://huggingface.co/bert-base-chinese}} model from Huggingface library \cite{wolf-etal-2020-transformers}. The token embedding layer is also initialized using BERT token embeddings. We fine-tune the model for 20 epochs. The parameters of the CNN-based feature encoder of wav2vec2.0 and the BERT model are permanently frozen, while the contextualized acoustic representation extractor of wav2vec2.0 is jointly updated from the $\text{5,001}^\text{th}$ training updates. The early stop strategy with patience 3 is used in our experiment to avoid overfitting. Finally, we average the best 10 checkpoints on the development set to obtain the final model. The batch size is set to 32 and the gradients are accumulated over 2 updates. We use the Adam optimizer with warm-up scheduler (25,000 steps) and the learning rate is $10^{-4}$ throughout all experiments.

\section{Experimental Results}
\label{sec:results}
\subsection{Results on AISHELL-1 dataset}

\begin{table}[t]
\centering
\begin{tabular}{m{5.5cm} M{0.7cm}M{0.7cm}} 
 \Xhline{1pt}
 Method&Dev&Test \\
 \hline \hline
 \textit{Autoregressive ASR} \\
 
 \quad Espnet2(w2v2) & 4.23 & 4.52 \\
 \quad Espnet2(w2v2) w/ LM & 4.20 & 4.47 \\ \hline
 
 \textit{Non-Autoregressive ASR} \\
 \quad Vanilla w2v2-CTC  & 4.85 & 5.13 \\
 \quad Espnet2(w2v2) w/ CTC-branch & 4.57 & 4.82 \\
 \quad KT-RL-ATT & 4.38 & 4.73 \\
%  \quad Proposed(no shift) & 4.27 & 4.52 \\
%  \quad Proposed(left) & 4.14 & 4.41 \\
 \quad CAKT w/ left-shift & \textbf{4.14} & \textbf{4.41} \\
 \quad CAKT w/ right-shift & \textbf{4.10} & \textbf{4.39} \\

\Xhline{1pt}
\end{tabular}
\caption{Results on AISHELL-1 dataset without external language model unless specified. }
\label{tab:aishell1_ours}
\end{table}

\begin{table}[]
    \centering
    \begin{tabular}{m{4.3cm}M{0.5cm}M{0.5cm}M{0.55cm}M{0.55cm}} 
    \Xhline{1pt}
 Method&PAM&PLM&Dev&Test \\
 \hline \hline
 \textit{Autoregressive ASR} \\
 \quad Espnet2(Trans.) \cite{watanabe2018espnet} w/ LM & & & 5.9 & 6.4 \\
 \quad Espnet2(Conf.) \cite{watanabe2018espnet} & & & 4.5 & 4.9 \\
 \quad Espnet2(Conf.) \cite{watanabe2018espnet} w/ LM & & & 4.4 & 4.7 \\
 \quad Preformer \cite{9688009} w/ LM & \checkmark & \checkmark & 4.3 & 4.6 \\
%  \quad Preformer w/ LM \cite{9688009} & \checkmark & \checkmark & 3.9 & 4.2 \\
 \hline
 \textit{Non-Autoregressive ASR} \\
 
 \quad LASO with BERT \cite{9437636} & & \checkmark & 5.2 & 5.8  \\
 \quad Vanilla w2v2-CTC \cite{9746316} & \checkmark & & 4.8 & 5.3 \\
 \quad KT-CL \cite{9747887} & \checkmark & \checkmark & 5.0 & 5.2 \\
 \quad KT-RL-ATT \cite{9747887} & \checkmark & \checkmark & 4.6 & 4.8 \\
 \quad rePLM-NAR-ASR \cite{9755057} & & \checkmark & 4.2 & 4.8 \\
 \quad KT-RL-CIF \cite{9747887} & \checkmark & \checkmark & 4.3 & 4.7 \\
 \quad NAR CTC/attention \cite{9746316} & \checkmark & \checkmark & 4.1 & 4.5 \\
 \quad Wav-BERT \cite{zheng-etal-2021-wav-bert} & \checkmark & \checkmark & 3.6 & 3.8  \\

% \quad CAKT & \checkmark & \checkmark & 4.10 & 4.39 \\
 
 \Xhline{1pt}
    \end{tabular}
    \caption{Results on AISHELL-1 dataset for various state-of-the-art systems. PAM and PLM indicate whether a pre-trained language model and/or a pre-trained speech representation learning method are/is used for building ASR.}
    \label{tab:aishell1_other}
\end{table}

\begin{table*}[]
    \centering
    \begin{tabular}{m{5cm} M{1.2cm} M{1.6cm} M{1.2cm} M{1.2cm} M{1.6cm} M{1.2cm}}
    \Xhline{1pt}
    \multirow{2}{*}{Method} & \multicolumn{3}{c}{Dev} & \multicolumn{3}{c}{Test} \\
    & iOS & Android & Mic & iOS & Android & Mic \\ 
    \hline \hline
    \textit{SOTA System} & & \\
        \quad Espnet1(Trans.) w/ LM \cite{watanabe2018espnet} & - & - & - & 7.5 & 8.9 & 8.6 \\
        \quad CTC-Enhanced(NAR) \cite{9414694} & - & - & - & 7.1 & 8.1 & 8.0 \\
        \quad CTC-Enhanced(AR) \cite{9414694} & - & - & - & 6.8 & 7.7 & 7.8 \\
        \quad LASO with BERT \cite{9437636} & 6.2 & 7.2 & 7.3 & 6.5 & 7.2 & 7.1 \\
        \quad rePLM-NAR-ASR \cite{9755057} & 5.5 & 6.1 & 6.2 & 5.7 & 6.3 & 6.2 \\
        \hline \hline
        \textit{Our Implementation} & & \\
        \quad Vanilla w2v2-CTC & 5.52 & 7.32 & 6.79 & 5.86 & 7.66 & 6.96 \\
        \quad KT-RL-ATT & 5.17 & 7.00 & 6.53 & 5.67 & 7.32 & 6.83 \\
        \quad CAKT w/ right-shift & 5.18 & 6.72 & 6.37 & 5.55 & 7.23 & 6.60 \\
    \Xhline{1pt}
    \end{tabular}
    \caption{Experimental results on AISHELL-2 dataset.}
    \label{tab:aishell2}
\end{table*}

In the first set of experiments, we evaluate the proposed CAKT and some baseline systems, which are implemented by ourselves. The experimental results are summarized in Table \ref{tab:aishell1_ours}. Espnet2(w2v2) is a hybrid CTC/attention-based model, whose encoder is wav2vec2.0 and the decoder consists of six layers of Transformer. The decoder has 4 heads and 2,048 hidden units. Since Espnet2(w2v2) is an autoregressive (AR) model, we can simply pair a language model with it for better results. The additional language model is also trained by the AISHELL-1 corpus. The vanilla w2v2-CTC and Espnet2(w2v2) with CTC-branch are classic non-autoregressive (NAR) models. The KT-RL-ATT concentrates on transferring knowledge from a pre-trained language model to ASR based on attention mechanism, so it can be treated as a strong baseline \cite{9747887}. Based on the results, the proposed CAKT surpasses all the baseline models. Specifically, CAKT with right-shift method yields a 14.4\% relative improvement over vanilla w2v2-CTC, and it is worth noting that their model size and decoding speed are identical at the inference phase. Furthermore, compared with autoregressive models, CAKT can achieve superior performances and provide a much faster decoding speed. Compared to the strong baseline system KT-RL-ATT, CAKT with right-shift method reduces the CER from 4.73\% to 4.39\% (i.e., 7.2\% relative improvement) showing a significant progress.

Next, we investigate the recognition performance for various state-of-the-art systems on the AISHELL-1 dataset. All the results are summarized in Table \ref{tab:aishell1_other}. Several worthwhile observations can be drawn. First, we divide all the ASR systems into AR and NAR models. The experimental results reveal that NAR models can obtain competitive results with AR models. Next, we take a step forward to analyze the contributions of using pre-trained language models and/or speech representation learning methods for ASR. Obviously, these pre-trained models can enhance the ASR performance for both AR and NAR models. Third, Preformer and Wav-BERT deliver the best results in autoregressive and non-autoregressive manners, respectively. It is noteworthy that both models combine a pre-trained language model and a pre-trained speech representation learning method to form the ASR model. As a result, they are not only larger but also more complicated than other methods.

Compared Table \ref{tab:aishell1_ours} to Table \ref{tab:aishell1_other}, the proposed CAKT demonstrates better results than most SOTA models. Therefore, we can conclude that the potential of CAKT comes from the token-dependent knowledge transferring mechanism and the context-aware training strategy. The former makes CAKT a knowledge-injected ASR, and the latter attends to ease the conditional independence assumption. Thereby, CAKT enjoys the benefits of a pre-trained speech representation learning method, mimics knowledge from a pre-trained language model, and is a lightweight ASR model with a fast decoding speed. In summary, CAKT presents an efficient and effective way to enhance CTC-based ASR.  

% Comparing with other benchmarks that utilize both pre-trained acoustic and language models, CAS-ASR shows comparable performance

% Second, comparing the results of KT-ATT-RL and Proposed(no shift), it shows that introducing discrete token embeddings to the transfer learning module provides better hints to extract representations from acoustic models more effectively. Furthermore, by learning representations from adjacency tokens, the misaligned auxiliary loss  improves the performance to the next level. The mechanism successfully helps CTC-based models to learn contextual information. Comparing the results from two shifted variants, the Proposed(right) has slightly better performance over Proposed(left). We consider it as dataset bias here.

\subsection{Results on AISHELL-2 dataset}
In addition to the AISHELL-1 dataset, we also carry out experiments on the industry-scaled AISHELL-2 corpus, and the results are shown in Table \ref{tab:aishell2}. The development and test sets of AISHELL-2 cover three different channels: iOS mobile phones (iOS), Android mobile phones (Android), and high-fidelity microphones (Mic). At first glance, all of the models can obtain better results on iOS than Andriod and Mic. Besides, the performance gaps between iOS and Android, as well as iOS and Mic, are much more extensive for wav2vec2.0-based models (i.e., vanilla w2v2-CTC, KT-RL-ATT, and CAKT). A possible reason is that the training data in AISHELL-2 are recorded by iOS devices, and the wav2vec2.0 is also pre-trained by using the training set. As such, these ASR models are much more accurate for iOS data than other data types. Next, the proposed CAKT yields better results than baseline systems, i.e., vanilla w2v2-CTC and KT-RL-ATT, in almost all cases. Compared with SOTA systems, CAKT also delivers better performance than most methods. Finally, although CAKT seems to obtain only comparative or worse results than rePLM-NAR-ASR, the model size of CAKT is much smaller than that of rePLM-NAR-ASR. Besides, CAKT is also faster than rePLM-NAR-ASR in the inference stage. Based on all the experiments, we can summarize that CAKT indeed achieve satisfactory performances on both AISHELL-1 and AISHELL-2 corpora. Furthermore, it is worth mentioning that the token-dependent knowledge transferring module and context-aware training strategy can be paired with other CTC-based ASR, and we believe these novel mechanisms can boost the progress of CTC-based models.

\label{sec:effectiveness}
\begin{table}[]
    \centering
    \begin{tabular}{M{4.5cm} M{1cm} M{0.7cm}M{0.7cm}}
    \Xhline{1pt}
    Query & Shift &  Dev & Test\\
    \hline \hline
                        Positional Embeddings & 0    & 4.39          & 4.68 \\
                        Positional Embeddings         & -1   & 4.36          & 4.60 \\
                        Positional Embeddings         & 1    & 4.36          & 4.66 \\
                      \hline
                      Token + Positional Embeddings  & 0    & 4.27          & 4.52 \\
                      Token + Positional Embeddings & -1   & 4.14          & 4.41 \\
                      Token + Positional Embeddings & 1    & \textbf{4.10} & \textbf{4.39} \\
    \Xhline{1pt}
    \end{tabular}
    \caption{Further analysis for the proposed CAKT ASR model on AISHELL-1 dataset.}
    \label{tab:module}
\end{table}

\subsection{Further Analysis}
Following, we turn to analyze the proposed CAKT, and the results are shown in Table \ref{tab:module}. First, we compare the efficiency of using token-dependent representations and positional embeddings as query vectors in the knowledge transferring step. The results indicate that token-dependent queries can obtain better results than positional embeddings. Although the change is simple, it can bring up to about 5.8\% relative improvements for both development and test sets. Next, we analyze the alignment methods used in the proposed context-aware training. When combined with positional embeddings, the right-shift and left-shift methods can only receive slight improvements than the no-shift method (i.e., Shift=0). However, when paired with token-dependent queries, the right-shift and left-shift methods deliver 2.8\% and 2.4\% relative improvements over the no-shift method on the test set, respectively. To sum up, the token-dependent queries play an important role in knowledge transferring and context-aware training.

\section{Conclusion}
This work presents a simple yet effective context-aware knowledge transferring strategy to inject linguistic knowledge and modulate the conditional independence assumption for CTC-based ASR. The knowledge transferring module is designed to distill knowledge from a pre-trained language model, and the context-aware training strategy forces the ASR model to be aware of contextual information. As such, the resulting CAKT model not only utilizes BERT and wav2vec2.0 but also relaxes the conditional independence assumption. Extensive experiments and analysis demonstrate the effectiveness and efficiency of the proposed method. We believe such a lightweight and simple method could be a potential basis for further research. In the future, we plan to evaluate the CAKT on other benchmark corpora and train it based on other pre-trained language models and speech representation learning methods. We will also continue improving the architecture and exploring different training objectives for the CTC-based ASR. 

\section{Acknowledgement}
This work was supported by the Ministry of Science and Technology of Taiwan under Grant MOST 111-2636-E-011-005 through the Young Scholar Fellowship Program. We thank the National Center for High-performance Computing (NCHC) of National Applied Research Laboratories (NARLabs) in Taiwan for providing computational and storage resources.

\bibliographystyle{IEEEbib}
\bibliography{strings,refs}

\end{CJK}
\end{document}